\documentclass[11pt]{article}
\usepackage[margin=1in]{geometry}
\usepackage{amsmath,amssymb,graphicx,booktabs,natbib,hyperref,xcolor}
\hypersetup{hidelinks} 
\newcommand{\todo}[1]{\textcolor{red}{[TODO: #1]}}
\newcommand{\sigstar}{\sigma^\star}       
\newcommand{\sigop}{\sigma_{\mathrm{op}}} 
\newcommand{\HarmPPtwentyfive}{$-0.092$}
\newcommand{\HarmPermuted}{$-0.190$}
\newcommand{\LadderDiffFash}{$+0.008\,[-0.004,+0.019]$}
\newcommand{\LadderDiffSplit}{$+0.024\,[+0.002,+0.052]$}
\newcommand{\LadderDoobLift}{$+0.095\,[+0.067,+0.121]$}
\newcommand{\LadderLinLift}{$+0.071\,[+0.046,+0.099]$}
\newcommand{\LadderOuLift}{$-0.011\,[-0.027,+0.003]$}
\newcommand{\LadderRandFixedFashLift}{$-0.101\,[-0.120,-0.082]$}
\newcommand{\LadderRandFixedSplitLift}{$-0.007\,[-0.016,+0.005]$}
\newcommand{\LadderRandStepLift}{$-0.018\,[-0.036,-0.000]$}
\newcommand{\LadderRatioFash}{$0.93\,[0.84,1.04]$}
\newcommand{\LadderRatioSplit}{$0.74\,[0.50,0.98]$}
\newcommand{\LengthAdvTeight}{$+0.303\,[+0.252,+0.342]$}
\newcommand{\LengthAdvTthirtytwo}{$+0.071\,[+0.028,+0.118]$}
\newcommand{\LengthRetSeq}{0.54/0.53/0.53}
\newcommand{\ScopeFashionLift}{$+0.110\,[+0.091,+0.127]$}
\newcommand{\ScopeMatchedLift}{$+0.009\,[+0.003,+0.015]$}
\newcommand{\ScopeMatchedRet}{0.63}
\newcommand{\ScopePermRet}{0.62}
\newcommand{\ScopePermSigStar}{0}
\newcommand{\TheoryRotSigHi}{0.02}
\newcommand{\TheoryRotSigLo}{0.005}
\newcommand{\TheoryRotSlope}{1.5}
\newcommand{\TheoryYYSigHi}{0.02}
\newcommand{\TheoryYYSigLo}{0.01}
\newcommand{\TheoryYYSlope}{0.5}

\title{A Noise Optimum in Rehearsal-Free Continual Learning:\\
Isolation, Mechanism, and Scope}
\author{Gunner Levi Howe\\ \href{mailto:gunnerlevihowe@gmail.com}{gunnerlevihowe@gmail.com}}
\date{}

\begin{document}
\maketitle

\begin{abstract}
Injecting stochastic noise into a consolidation rule can \emph{improve} a network's
retention of earlier tasks up to an optimal level, then degrade it --- an inverted-U in
retention vs.\ noise. This paper isolates what produces that optimum and maps where it
holds, entirely in simulation. (1) \textbf{Phenomenon:} the retention inverted-U appears
on several related-task continual-learning benchmarks (Split-MNIST, FashionMNIST,
continual Yin-Yang). (2) \textbf{Isolation:} a magnitude-matched ladder shows the effect
requires \emph{coherent restoring toward the consolidated weights} --- a random-direction
force of identical magnitude produces no optimum, and a coherent force toward the wrong
target actively hurts. (3) \textbf{Active ingredient:} most of the optimum is recovered
by coupling the anchor gain to the \emph{injected-noise variance} $\sigma^2$ --- a
one-line rule that neither Ornstein-Uhlenbeck Adaptation (fixed gain) nor MESU
(posterior-variance gain) implements. A forced Ornstein-Uhlenbeck calculation derives the
rising flank and predicts that the optimal noise rises with per-task interference $g$ ---
confirmed out-of-sample \emph{in direction} against pre-existing measurements (the exponent
is unresolved at our grid). The barrier-conditioning of the
originating Doob $h$-transform is a low-$\sigma$ safety net that bounds forgetting where
the coupled gain is too weak. (4) \textbf{Scope:} the optimum requires shared task
structure --- it is absent on permuted-MNIST, and a controlled
rotated-vs-permuted comparison localizes the boundary to task structure; the precise
governing quantity is left open. (5) \textbf{Length:} at matched severity the advantage
persists but attenuates with task count, and we show no rotation family can attribute
the trend (a compact-group identity). A single-seed BrainScaleS-2 demonstration of the
originating rule is reported separately \citep{paper1}; this paper makes no hardware claim.
\end{abstract}

\section{Introduction}
Continual learning on a single network suffers catastrophic forgetting: training on a new
task overwrites the weights that encoded earlier ones. Rehearsal-free consolidation methods
resist this by anchoring important weights, but they treat stochastic noise as a nuisance to
be minimized. This paper studies the opposite regime, in simulation: a consolidation rule
into which injected noise is deliberately coupled, and which \emph{improves} retention up to
an optimal noise level, then degrades it --- an inverted-U in retention versus noise. A
companion paper \citep{paper1} introduced the rule (a Doob barrier-conditioned diffusion) and
demonstrated it on BrainScaleS-2 analog hardware, where the noise is intrinsic and free; here
we ask, entirely digitally, \emph{what} produces the optimum, \emph{where} it holds, and
\emph{how} it behaves with task count. Our contributions: (i) a magnitude-matched
\emph{ladder} that isolates the active ingredient --- coherent restoring toward the
consolidated weights, with most of the effect carried by a one-line coupling of the anchor
gain to the injected-noise variance, distinct from the fixed-gain (OUA) and posterior-variance
(MESU) baselines; (ii) a forced-Ornstein-Uhlenbeck account that derives the rising flank and
predicts the optimum rises with per-task interference (direction confirmed out-of-sample);
(iii) a scope law --- the optimum requires shared task structure, with the scalarizable
mediator left honestly open; and (iv) a length characterization with a compact-group
identity showing the trend is unattributable on any rotation family.

\section{The retention inverted-U}\label{sec:phenom}
On the primary Split-MNIST domain-incremental stream, coupling the injected noise into the
consolidation rule produces an inverted-U in retention: retention rises to an interior
optimum near $\sigma=0.02$ (lift over zero noise \LadderDoobLift) and falls thereafter,
while the matched anchored-drift control is flat-to-monotone in noise (\LadderOuLift\ at
$\sigop$; the full monotone sweep is in the companion paper \citep{paper1}). The
optimum reproduces on FashionMNIST (\ScopeFashionLift) and on the procedural continual
Yin-Yang stream. Throughout we report bootstrap confidence intervals rather than $p$-values:
a one-sided Wilcoxon signed-rank test at $n$ seeds floors at $2^{-n}$, so at our seed counts
a ``significant'' $p$ is exactly ``all seeds agreed'' and carries no effect-size information.

\section{Isolating the active ingredient: a steering-force ladder}\label{sec:ladder}
The $\kappa\!:\!1\!\to\!0$ ablation of the companion paper \citep{paper1} deletes the entire steering term and so cannot say
\emph{which part} of it carries the inverted-U. Holding the anchor pull and the injected
noise identical, we replace the steering score with magnitude-matched controls (a one-line
hook): \emph{random-step} (the score's per-weight magnitude, sign resampled every step ---
incoherent); \emph{random-fixed} (magnitude, sign fixed per run --- coherent toward a
\emph{wrong} target); \emph{linear} (a $\sigma^2$-coupled linear restoring toward the
anchor, the barrier divergence removed); and full \emph{Doob}. Lift is read at
$\sigma=0.02$, the operating point fixed on separate prior seeds (not selected
per-condition on this run, so no winner's-curse inflation), with bootstrap CIs, on
Split-MNIST and FashionMNIST ($n{=}8$ seeds).

\paragraph{Coherent restoring toward memory is required.} Neither random rung yields an
inverted-U: random-step sits at the unconditioned baseline (\LadderRandStepLift), and
random-fixed --- a coherent force toward the wrong target --- is near-neutral on Split-MNIST
(\LadderRandFixedSplitLift) but actively \emph{degrades} retention on FashionMNIST
(\LadderRandFixedFashLift, worse than injecting nothing); this Split/Fashion asymmetry in the
wrong-target rung we do not explain. The force must point at the consolidated weights; matched
magnitude and mere coherence are insufficient.

\paragraph{The active ingredient is the noise-coupled gain, not the barrier.} The linear
rung recovers most of the optimum (lift \LadderLinLift\ vs.\ Doob's \LadderDoobLift\ on
Split-MNIST) --- linear/Doob lift ratio \LadderRatioSplit\ on Split-MNIST and
\LadderRatioFash\ on FashionMNIST. Crucially this rung is \emph{not} the
surrendered anchored drift: OUA's mean-reversion gain is independent of the noise
(Eq.~5 of \citealp{oua2024}), and MESU scales its anchor pull by the \emph{posterior}
variance, not the injected noise (Eq.~11 of \citealp{mesu2025}). Coupling the consolidation
gain to the \emph{injected-noise} variance $\sigma^2$ is what produces the optimum, and it
is a one-line modification neither method makes.

\begin{figure}[t]\centering
\includegraphics[width=\textwidth]{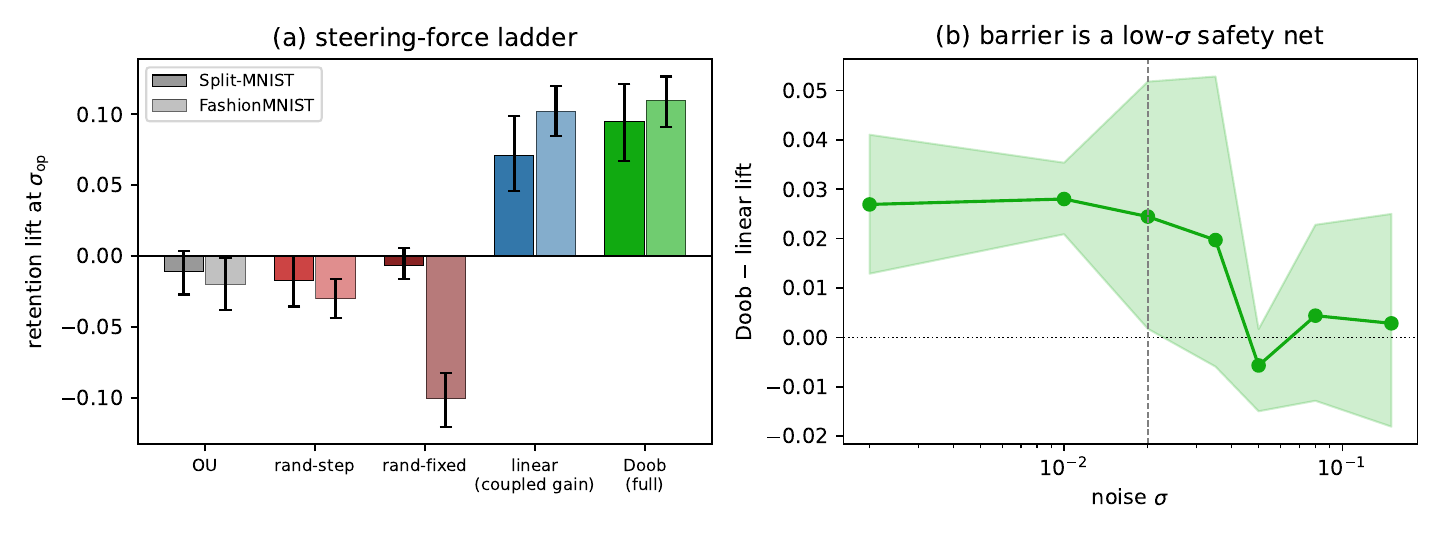}
\caption{\textbf{Steering-force ladder.} (a) Retention lift at $\sigop$ for each rung
(Split-MNIST solid, FashionMNIST faded), bootstrap CIs: random-direction forces (incoherent
or wrong-target) give no lift, while the $\sigma^2$-coupled linear rung recovers most of the
Doob lift. (b) The Doob$-$linear difference across noise --- a significant low-$\sigma$
increment that vanishes at high $\sigma$: the barrier is a low-$\sigma$ safety net
(\S\ref{sec:theory}).}
\label{fig:ladder}
\end{figure}

\paragraph{The barrier is a low-$\sigma$ safety net.} Beyond the linear rung, the full Doob
barrier adds \LadderDiffSplit\ on Split-MNIST (significant) and \LadderDiffFash\ on
FashionMNIST (not). This increment is significant and roughly constant at low noise
($\sigma\!\le\!0.02$) and vanishes by $\sigma\!\ge\!0.05$ (Fig.~\ref{fig:ladder}), the
signature the forced-OU model predicts (\S\ref{sec:theory}): the barrier bounds forgetting
in the low-noise regime where the coupled gain alone is too weak, and is idle at high
noise. Its benchmark-specificity is consistent with the engagement criterion
$g/(s\sigma^2)\gtrsim b$ (stated as consistent, not verified).

\section{Theory: a forced Ornstein-Uhlenbeck account}\label{sec:theory}
Model a consolidated weight under new-task interference as a forced OU process,
$dw=(-\theta w + g)\,dt + \sigma\,dW$, with $g$ the time-averaged interference pressure
pushing $w$ from the anchor and $\theta$ the restoring gain. The stationary displacement is
$g/\theta$ and the spread $\sigma^2/2\theta$.
\begin{itemize}
\item \textbf{Fixed gain} $\theta=s$: displacement $g/s$ (noise-independent), spread
$\sigma^2/2s$ --- retention loss \emph{grows} with noise, deriving the monotone-decreasing
baselines.
\item \textbf{Noise-coupled gain} $\theta=s\sigma^2$: displacement $g/(s\sigma^2)$, spread
$1/2s$ --- loss $g^2/(s^2\sigma^4)+1/2s$ \emph{falls} with noise: the inverted-U's rising
flank. (The falling flank we posit as a high-$\sigma$ plasticity tax; $g$ is an effective
time-averaged pressure --- both flagged as modeling assumptions.)
\end{itemize}
If the high-$\sigma$ plasticity tax grows as $c\sigma^2$, minimizing
$g^2/(s^2\sigma^4)+c\sigma^2$ gives an optimum $\sigstar\!\propto\!g^{1/3}$; we treat both the
tax form and $g\!\propto\!$ interference as modeling assumptions and test only what the data
bear. \emph{Out-of-sample} the \emph{direction} is confirmed --- $\sigstar$ rises with
per-task interference (rotated-MNIST $10^\circ\!\to\!25^\circ$:
$\sigstar{=}\TheoryRotSigLo\!\to\!\TheoryRotSigHi$; Yin-Yang length scan:
$\sigstar{=}\TheoryYYSigHi\!\to\!\TheoryYYSigLo$ as step size falls) --- but the
\emph{exponent} is not: the implied slopes ($\approx\!\TheoryRotSlope$ and $\approx\!\TheoryYYSlope$) are unresolved
at our $\times2$ noise grid, and the Yin-Yang pair inherits the step/length confound of
\S\ref{sec:length}. The barrier's
low-$\sigma$ role follows: the coupled displacement $g/(s\sigma^2)$ is largest exactly where
the gain is weakest, so at low $\sigma$ the coupled rule alone lets interference drag weights
past the barrier while the divergence bounds them (\S\ref{sec:ladder}).

\section{Scope: the optimum requires shared task structure}\label{sec:scope}
The optimum does not hold universally. On permuted-MNIST --- where a fixed random pixel
permutation per task destroys the shared image manifold --- there is no inverted-U (at the
tested task count $T{=}10$): the retention optimum sits at zero noise
($\sigstar=\ScopePermSigStar$), i.e.\ noise only hurts. A controlled contrast isolates the
boundary to task \emph{structure} rather than forgetting \emph{severity}. Holding
architecture, operating point and task count fixed and varying only the per-task input
transform, rotated-MNIST (which preserves the manifold) shows the inverted-U while
permuted-MNIST does not. Decisively, a structured cell at permuted's \emph{own} mild severity
(rotated $8^\circ$, retention $\ScopeMatchedRet$ vs.\ permuted's $\ScopePermRet$) still shows a
significant optimum (lift \ScopeMatchedLift, $16$ seeds) while permuted at that severity
shows none --- so at matched forgetting it is task structure, not severity, that gates the
optimum. A
partial-permutation dial (shuffling $k\%$ of pixels per task) shows no optimum at any $k$;
but along this dial shared pixels both constitute the structure and suppress the interference,
so every cell sits in the mild, low-forgetting regime where no benchmark of any kind shows the
optimum: the dial is a confounded instrument and cannot by itself exclude a pixel-fraction
account. What it \emph{does} show cleanly is a monotone \emph{harm} gradient at the operating
noise, from \HarmPPtwentyfive\ at $k{=}25\%$ (most structure) to \HarmPermuted\ at full
permutation --- the noise-tax-plus-displacement-drag that is the model's negative branch. We also measured a candidate scalar,
init-relative consecutive-task solution alignment computed layerwise, and found it does not
separate the families at any layer --- init-relative weight deltas track the discriminative
gradient short training writes, not the input manifold, so different digit pairs look as
unaligned as scrambled inputs. We therefore state the scope empirically: the optimum
requires shared task structure, and we leave the scalarizable mediator open, naming two
untested constructs for future work: the \emph{sequential} weight displacement along the
trained trajectory (which our init-relative alignment does not capture), and input-statistics
overlap (task-input covariance / centered kernel alignment).

\begin{figure}[t]\centering
\includegraphics[width=0.62\textwidth]{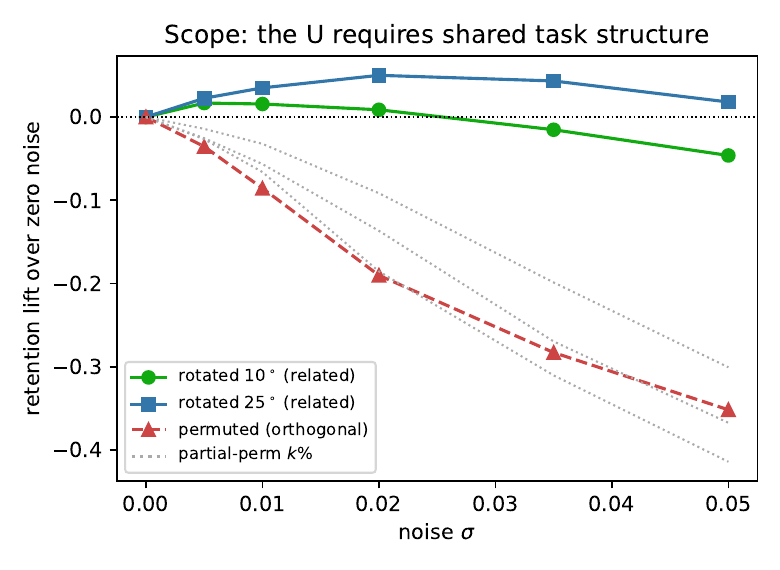}
\caption{\textbf{Scope.} Retention lift vs.\ noise: the inverted-U appears on rotated-MNIST
(shared manifold) and is absent on permuted-MNIST and on the partial-permutation dial at
every $k$. Shared task structure, not noise magnitude or pixel-overlap, gates the optimum.}
\label{fig:scope}
\end{figure}

\section{Length: attenuation, and why no rotation family can attribute it}\label{sec:length}
At matched severity the advantage over the control attenuates with task count. Using
continual Yin-Yang with the total rotation span held fixed at $180^\circ$ (so the per-task
step, not the severity, absorbs the change in $T$), the normalized doob$-$OU advantage is
significant at every tested length --- \LengthAdvTeight\ at $T{=}8$ falling to
\LengthAdvTthirtytwo\ at $T{=}32$ ($16$ seeds, $\sigma$ selected on an $8$/$8$ split) --- with
zero-noise retention held constant across $T$ ($\LengthRetSeq$, confirming matched severity).
We do not attribute the decline. On any rotation family the span equals step$\times(T-1)$
and severity is a function of span: two constraints on one free knob, so length co-varies
exactly with per-task step and no rotation design can separate them. (The apparent
\emph{growth} at short $T$ under a fixed step and the \emph{decay} here under a fixed span
are both artifacts of which knob is held.) The robust claim is only that the advantage
persists and remains significant through $T{=}32$; attributing the trend requires a
constant-per-task-novelty class sequence such as Split-CIFAR-100, which we flag as in
progress.

\begin{figure}[t]\centering
\includegraphics[width=0.62\textwidth]{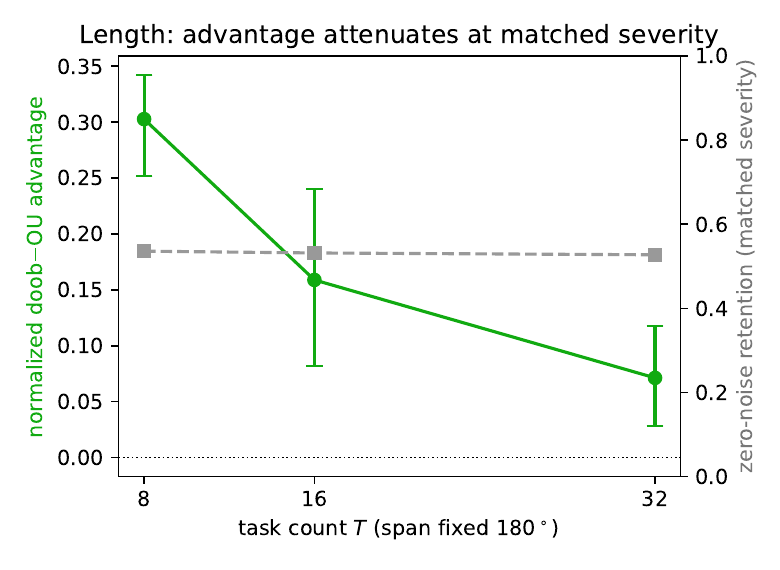}
\caption{\textbf{Length.} Total rotation span held fixed at $180^\circ$: zero-noise retention
is constant across task count (matched severity) while the normalized advantage attenuates.
The trend is real but, by the compact-group identity (\S\ref{sec:length}), unattributable on
any rotation family.}
\label{fig:length}
\end{figure}

\section{What we do not claim}\label{sec:noclaim}
\begin{itemize}
\item \textbf{Not a retention state-of-the-art.} Rehearsal, which stores past data,
out-retains this rehearsal-free rule; among rehearsal-free methods it is competitive, not
dominant. The contribution is the mechanism and its scope, not a leaderboard entry.
\item \textbf{No hardware claim.} This paper is entirely in simulation. The single-seed
BrainScaleS-2 demonstration of the originating rule is reported separately \citep{paper1};
we make no silicon claim here.
\item \textbf{The Doob $h$-transform is not the load-bearing mechanism.} The ladder
(\S\ref{sec:ladder}) attributes most of the optimum to the noise-coupled gain; the barrier
divergence is a low-noise refinement. Doob is the derivation route that produced the
coupling, not the effect itself.
\item \textbf{We do not identify the scope mediator} (\S\ref{sec:scope}), and \textbf{we do
not attribute the length trend} (\S\ref{sec:length}) with the instruments used here.
\end{itemize}

\section{Reproducibility}
All experiments run on CPU with fixed seeds. Every result number in the prose is generated
into \texttt{numbers.tex} by \texttt{gen\_paper\_numbers.py} from committed
\texttt{results/*.json}; \texttt{verify\_regen.py} enforces byte-identity, so no result number
is hand-typed. The fixed operating point (task/consolidation learning rate $0.1$, barrier
scale $0.2$, operating noise $\sigop=0.02$), the seed counts, and the noise grids are stated with each
experiment. Code and data: \todo{repo URL / archival DOI at submission}.

\bibliographystyle{plainnat}
\bibliography{references}
\end{document}